\documentclass[letterpaper,10pt,conference]{IEEEtran}
\IEEEoverridecommandlockouts
\usepackage{cite}
\usepackage{amsmath,amssymb,amsfonts}
\usepackage[ruled,noend,linesnumbered]{algorithm2e}
\usepackage{booktabs}
\usepackage{tabularx}
\usepackage[breaklinks,hypertexnames=false]{hyperref}
\usepackage{graphicx}
\usepackage{textcomp}
\usepackage[nameinlink]{cleveref}
\crefname{figure}{Fig.}{Figs.}\Crefname{figure}{Fig.}{Figs.}
\crefname{table}{Table}{Tables}\Crefname{table}{Table}{Tables}
\crefname{section}{Section}{Sections}\Crefname{section}{Section}{Sections}
\crefname{subsection}{Section}{Sections}\Crefname{subsection}{Section}{Sections}
\crefname{algocf}{Algorithm}{Algorithms}\Crefname{algocf}{Algorithm}{Algorithms}
\newcommand{\lcref}[1]{\hyperref[#1]{line~\ref*{#1}}}
\newcommand{\Lcref}[1]{\hyperref[#1]{Line~\ref*{#1}}}
\usepackage[all]{hypcap}
\hypersetup{
    colorlinks,
    linkcolor={red!50!black},
    citecolor={blue!50!black},
    urlcolor={blue!80!black}
}

\usepackage[colorinlistoftodos]{todonotes}
\usepackage{xcolor}
\usepackage[most]{tcolorbox}
\usepackage{color,soul}
\usepackage{tikz}
\usepackage[switch]{lineno}

\usepackage{scalerel}
\usetikzlibrary{svg.path}

\title{Efficient Constant Optimization for Symbolic Regression with GPU-Accelerated Tree-Based Genetic Programming}

\author{
\IEEEauthorblockN{
    Hao Mao\IEEEauthorrefmark{1}
    Xu T. Liu\IEEEauthorrefmark{2}\IEEEauthorrefmark{3}
    Shuai Lu\IEEEauthorrefmark{4}
    Peng Zhao\IEEEauthorrefmark{5}
    Wenzheng Jiang\IEEEauthorrefmark{6}
    Yuntian Chen\IEEEauthorrefmark{7}
    }
    \IEEEauthorblockA{
    \IEEEauthorrefmark{1}The Hong Kong Polytechnic University, Hong Kong SAR, China
    \IEEEauthorrefmark{2}University of Washington, USA \\
    \IEEEauthorrefmark{3}Amazon Web Services, USA  
    \IEEEauthorrefmark{4}Jiangxi University of Finance and Economics, China
    \IEEEauthorrefmark{5}EEO Education Technology, China \\
    \IEEEauthorrefmark{6}Chongqing Medical University, China
    \IEEEauthorrefmark{7}Eastern Institute of Technology, Ningbo, China 
    }
}

\begin{document}

\maketitle

\begin{abstract}
Constant optimization refines the numerical coefficients of candidate expressions in tree-based genetic programming for symbolic regression. But its per-generation cost has led modern GPU-accelerated frameworks to omit it or restrict it to lightweight forms. We present a GPU-resident, batched Levenberg--Marquardt solver that optimizes constants across a structurally heterogeneous population of expression trees using a fixed number of population-wide CUDA launches per iteration. Reverse-mode automatic differentiation assembles the per-tree Jacobian in one backward sweep, making the dominant per-iteration cost independent of the number of constants per tree, and a double-precision delivery guard guarantees that returned constants are never worse than their initial values. On early-generation populations, the solver sustains up to $5.1{\times}10^{5}$ trees per second on an NVIDIA A100; at a GPU-saturated benchmark configuration it delivers roughly $9.9{\times}$ the throughput of Operon running on a 64-core EPYC 7763, while matching fp64-reference quality. Integrated in-process into EvoGP, the solver enables end-to-end search to recover governing equations on $10$ of $18$ constructed problems versus 0 for stock EvoGP. 
Our code is at https://github.com/TensorConv/CuSR.

\end{abstract}

\section{Introduction}
Symbolic regression (SR) discovers closed-form expressions from
data by jointly searching over both the structure and the numerical parameters
of candidate models. The dominant approach is tree-based genetic programming
(TGP), which evolves a population of expression trees through fitness-based
selection and genetic operators such as subtree crossover and mutation.
However, genetic operators modify tree topology and are ineffective at
fine-tuning real-valued constants; consequently, a structurally correct
expression may receive poor fitness because its numerical coefficients are
suboptimal. Constant optimization (CO) addresses this by solving,
for each fixed tree, a continuous nonlinear least-squares
subproblem via local optimization, typically Levenberg--Marquardt (LM).

The tension between the benefit of CO and its computational cost is reflected in the design of modern SR systems. PySR, a Julia-based framework on CPU, applies BFGS optimization with a restricted iteration budget per generation to control overhead~\cite{Cranmer2023PySR}. Operon, a high-performance C++ library, integrates LM constant optimization but operates within CPU-bound throughput constraints~\cite{Burlacu2020Operon}. EvoGP, a GPU-based TGP framework, achieves population-level parallelism through tensorized tree representations and GPU-resident evolutionary operators, yielding substantial gains in evaluation throughput~\cite{EvoGP_TEVC_2026}.
Critically, however, EvoGP does not natively support constant optimization, leaving a gap between its structural search and the precision needed to recover governing equations with inner constants: coefficients that sit inside a nonlinear function.

Closing this gap is not a straightforward port of existing CPU solvers. In a GPU-resident evolutionary loop, off-GPU constant optimization would incur repeated host--device data transfers that negate the throughput advantage of GPU acceleration. The workload is further complicated by structural heterogeneity: different expression trees contain different numbers of constants and may require different numbers of optimization iterations, which is incompatible with the uniform batched execution model that GPU architectures demand.

In this paper, we address these challenges by designing a GPU-resident, batched LM solver purpose-built for the heterogeneous CO workload in TGP-based symbolic regression, and integrating it directly into EvoGP so that constant optimization runs entirely on the GPU within each generation.
Our method co-designs the optimization kernels with the tree evaluation pipeline. A reverse-mode automatic differentiation pass builds the per-tree Jacobian in a single backward sweep.
The LM loop executes as a fixed number of population-wide CUDA kernel launches per iteration, and a double-precision delivery guard certifies that the resulting constants are never worse than their initial values.
In end-to-end searches on problems constructed to require nonlinear inner-constant fitting, EvoGP with in-loop constant optimization recovers the true governing equations that stock EvoGP does not. To the best of our knowledge, this is the first work to bring batched, second-order constant optimization fully onto the GPU for tree-based symbolic regression.
Below are our main contributions.
\begin{enumerate}
\item We present a GPU-resident batched second-order LM primitive that optimizes the constants of a structurally heterogeneous population in a fixed number of population-wide CUDA launches per iteration; reverse-mode automatic differentiation keeps the dominant per-iteration cost independent of per-tree constant count, and a double-precision delivery guard certifies that constants are never worse than their initial values (Section~\ref{sec:batched}).
\item We integrate this primitive in-process into EvoGP, eliminating per-generation subprocess launches and CUDA-context rebuilds so that constant optimization runs inside the search loop; applying it every fifth generation is statistically indistinguishable from applying it every generation (Sections~\ref{sec:integration} and~\ref{subsec:operon}).
\item We build a controlled benchmark for constant optimization (heterogeneous populations calibrated to real GP runs, with known near-optimal targets, solved identically on CPU and GPU), and on constructed inner-constant problems we show that the in-loop GPU solver recovers governing equations that structural evolution alone does not (Sections~\ref{subsec:workload} and~\ref{subsec:operon}).
\end{enumerate}

\section{Background and Related Works}

We introduce the notation used throughout and review prior work on tree-based genetic programming, constant optimization, and computational methods for symbolic regression.

\subsection{Tree-Based GP and Constant Optimization}

The dataset for symbolic regression is denoted as $\mathcal{D}=\{(\mathbf{x}_{i},y_{i})\}_{i=1}^{N}$, where $\mathbf{x}_{i}\in\mathbb{R}^{d}$ is the input vector, $y_{i}\in\mathbb{R}$ is the target value, $N$ is the number of samples, and $d$ is the number of input variables. An expression is represented by a tree $T$, where internal nodes are functions or operators and leaf nodes are variables or constants. The function set is denoted as $\mathcal{F}$, the terminal set is denoted as $\mathcal{T}$, and the set of real-valued constants in $T$ is denoted as $\mathbf{c}_{T}\in\mathbb{R}^{K_{T}}$, where $K_{T}$ is the number of constants in the tree.

For a given tree $T$, the expression evaluated on an input vector $\mathbf{x}$ is written as $f_{T}(\mathbf{x};\mathbf{c}_{T})$. The symbolic regression objective considered in this paper is to find both a tree structure and its constants by minimizing the prediction loss:
\begin{equation}
\min_{T,\mathbf{c}_{T}} \; \mathcal{L}(T,\mathbf{c}_{T})=
\frac{1}{N}\sum_{i=1}^{N}\left(f_{T}(\mathbf{x}_{i};\mathbf{c}_{T})-y_{i}\right)^{2}.
\label{eq:sr_objective}
\end{equation}
The structure $T$ is discrete and changes through genetic programming operators, while $\mathbf{c}_{T}$ is continuous and can be refined by numerical optimization. In a population-based algorithm, the population at generation $g$ is denoted as $\mathcal{P}^{(g)}=\{T_{1}^{(g)},T_{2}^{(g)},\ldots,T_{M}^{(g)}\}$, where $M$ is the population size.

Tree-based genetic programming (TGP) is a common representation for symbolic regression. Each individual in the population is an expression tree~\cite{Makke2024SRReview,Orzechowski2018Benchmark}. During evolution, individuals are selected according to fitness, and new individuals are generated by genetic operators such as subtree crossover, subtree mutation, point mutation, and reproduction. Since a tree directly corresponds to a mathematical expression, TGP can search over flexible nonlinear structures and produce readable analytic forms.

Compared with regression methods that assume a fixed model family, TGP searches over model structure and model size. This flexibility makes it suitable for scientific discovery and interpretable machine learning, where the expression is expected to be compact and meaningful rather than only accurate~\cite{Udrescu2020AIFeynman,Makke2024SRReview}. Closely related is the data-driven discovery of governing differential equations, where symbolic and evolutionary methods search over equation structure while fitting its coefficients~\cite{Chen2022SGAPDE,Xu2020DLGAPDE,Lou2026DEDiscoverySurvey}. However, TGP also faces two computational difficulties. First, fitness evaluation is expensive because each candidate expression must be evaluated on all training samples. Second, evolutionary operators mainly change tree structure and are not efficient local optimizers for real-valued constants. Therefore, a structurally promising tree may receive a poor fitness value if its constants are inaccurate~\cite{Orzechowski2018Benchmark,Kommenda2013CO}.

To address this issue, constant optimization is often introduced into GP-based symbolic regression. For a fixed tree structure $T$, constant optimization solves:
\begin{equation}
\mathbf{c}_{T}^{*}=\mathop{\arg\min}_{\mathbf{c}_{T}}
\frac{1}{N}\sum_{i=1}^{N}\left(f_{T}(\mathbf{x}_{i};\mathbf{c}_{T})-y_{i}\right)^{2}.
\label{eq:constant_opt}
\end{equation}
Previous studies have shown that numerical optimization methods, such as nonlinear least-squares optimization and evolutionary strategies, can improve fitted expressions and provide a more informative fitness signal for selection~\cite{Kommenda2013CO,Kommenda2020NLS,Alonso2009ESConstants}. At the same time, constant optimization is not a trivial add-on. Applying local optimization to every individual can introduce overhead, especially when trees have different structures and constant counts~\cite{Kommenda2020NLS}; it is also often ill-conditioned~\cite{Kronberger2022IllConditioned}.

\subsection{Computational Approaches to Symbolic Regression}

Traditional symbolic regression systems are implemented on CPUs. CPU-based implementations provide flexible control flow and can use mature numerical optimization libraries, making them suitable for evaluating expression trees and applying individual-specific constant optimization~\cite{Cranmer2023PySR,Burlacu2020Operon}. However, evaluating a population over many training samples remains computationally expensive, even when multithreading or vectorized execution is used~\cite{Chitty2012CPUvsGPU,Baeta2021SpeedBenchmark}.

The main cost of TGP-based symbolic regression comes from evaluating expression trees over many samples. This pattern provides parallelism, making GPU acceleration attractive. GPU-based GP and symbolic regression systems accelerate program evaluation and population-level fitness computation through CUDA-style parallel execution or GPU-resident workflows~\cite{Langdon2010CUDA,Langdon2011GPUOverview,Baeta2021TensorGP,SRgpu_pricai_2022,EvoGP_TEVC_2026}. These systems show that high performance depends not only on the GPU, but also on data layouts and evaluation kernels that reduce irregular memory access and avoid CPU--GPU synchronization~\cite{Lu2023Im2winGPU}.

Most GPU-accelerated TGP systems focus on structural evolution and fast fitness evaluation~\cite{SRgpu_pricai_2022,EvoGP_TEVC_2026}. Constant optimization is a different workload: calling a local optimizer per expression, as CPU systems do, fits the GPU poorly, since trees carry different numbers of constants and repeated host--device transfer erodes the benefit of GPU acceleration. Constant optimization for GPU-based TGP therefore needs a batched, population-compatible design.

The most related GPU-based symbolic regression system is Kozax~\cite{deVries2025Kozax}, a JAX-based genetic-programming framework that supports numerical constant fitting. However, its constant optimization is an optional, first-order component. In contrast, this paper focuses on constant optimization itself. We design a GPU-compatible second-order constant-optimization kernel that refines constants for heterogeneous expression trees while preserving the throughput advantage of GPU-accelerated tree-based genetic programming.

\begin{figure*}[t]
\centering
\includegraphics[width=0.92\textwidth]{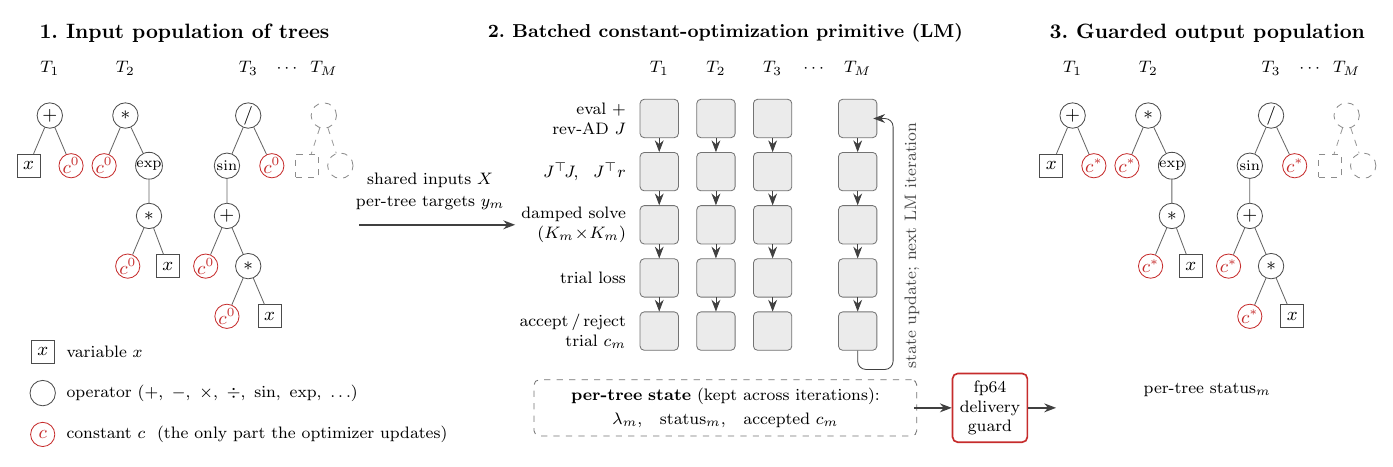}
\caption{The batched constant-optimization primitive. Left: a population of
$M$ heterogeneous expression trees, each with its own initial constants
$\mathbf{c}^{0}_m$ (red). Middle: a batched Levenberg--Marquardt loop
advances all trees together. Right: a double-precision delivery guard runs
before the fitted constants and per-tree status are returned.}
\label{fig:overview}
\end{figure*}

\section{Batched Constant Optimization for Heterogeneous Trees}
\label{sec:batched}

This section develops a GPU-resident, batched LM primitive that fits constants for an entire population of trees at once. We motivate the design first, then build it up from its contract to its deployment inside EvoGP.

\subsection{Motivation}
\label{subsec:motivation}

EvoGP touches constants only through random initialization and mutation. A
tree with the right structure can still lose: unfit constants keep its loss
high, and selection discards it for a bloated approximation, not the true
formula. Fitting constants is solved for one tree. It is not solved for a
population, every generation. Our goal is to make constant optimization a
batched GPU primitive, cheap enough to run for every tree in every generation of the search.

\subsection{Proposed Constant-Optimization Workflow}
\label{subsec:workflow}

Here is our proposed constant-optimization workflow, as illustrated in~\Cref{fig:overview}.
We treat the population, not the single tree, as the unit of constant
optimization input: one call to a GPU-resident primitive fits the constants of
all $M$ trees.
In \Cref{fig:overview}, the primitive takes a population
of $M$ expression trees $T_1,\dots,T_M$, a per-tree vector of initial
constants $\mathbf{c}^{0}_m\in\mathbb{R}^{K_m}$, where $m\in\{1,\dots,M\}$ and $K_m$ is the constant count of $T_m$, and a per-tree target $\mathbf{y}_m$ over a shared set of $N$ input
points $X$.
The population shares only these input points $X$, not initial constants or target $\mathbf{y}_m$.
Once the solve is underway, each tree also converges at its own pace.

Each call to the primitive runs in three stages. It first encodes the trees
into a structure-of-arrays batch layout: per-node and per-tree fields are
kept in separate, contiguous arrays across the whole population. Second, a batched
LM loop advances all trees together: each iteration
builds the Jacobian of every tree, solves its damped normal equations,
accepts or rejects its trial step. Third, before returning, a
double-precision delivery guard re-evaluates the loss of every tree. The output
is the fitted constants and a per-tree status: converged, failed, or capped
at the iteration limit. It also has one guarantee: evaluated in double precision (fp64),
every tree's delivered constants are never worse than its initial ones. Any
tree whose fit ends worse or non-finite is rolled back to its initial
values. Trees without constants ($K_m=0$) pass through untouched, and the
tree structure itself is never modified, only its constants.

\subsection{GPU-Batched Levenberg--Marquardt}
\label{subsec:lm}

We present the GPU-batched LM loop as~\Cref{alg:lm}.
\begin{algorithm}[t]
\small

\SetKwInOut{Input}{In}\SetKwInOut{Output}{Out}
\SetKwFor{ParFor}{parallel for}{do}{}
\SetKwFor{ParForR}{parallel for}{do}{}
\caption{GPU-batched Levenberg--Marquardt}
\label{alg:lm}
\Input{node/const.\ buffers; $\mathbf{c}^{0}_m$; $X,\{\mathbf{y}_m\}$; $\lambda_{0}$; $t_{\max}$}
\Output{constants $\mathbf{c}_m$ and status$_m$, $m\in\{1,\dots,M\}$}
$\mathbf{c}_m\leftarrow\mathbf{c}^{0}_m$;\ $\lambda_m\leftarrow\lambda_{0}$;\ status$_m\leftarrow$ active ($K_m{=}0$: skip)\;
$\ell^{\star}_m\leftarrow\mathrm{loss}(\mathbf{c}_m)$\tcp*{batched eval; fills $\mathbf{r}_m$}
\For(\tcp*[f]{outer loop on host}){$t\leftarrow1$ \KwTo $t_{\max}$\nllabel{alg:lm:line:outer}}{
  \lIf{no tree is active}{\textbf{break}}
  \ParFor{active tree $m\in\{1,\dots,M\}$}{
    \ParForR{data row $i\in\{1,\dots,N\}$}{
      $J_m[i,1{:}K_m]\leftarrow\textsc{BuildJacobianRow}(T_m,\mathbf{c}_m,\mathbf{x}_i)$\nllabel{alg:lm:line:jacobian}\;
    }
  }
  \ParFor{active tree $m$}{
    $A_m\leftarrow J_m^{\top}J_m$;\ \ $\mathbf{g}_m\leftarrow J_m^{\top}\mathbf{r}_m$\tcp*{reduce over rows}
  }
  \ParFor{active tree $m$}{
    solve $\bigl(A_m+\lambda_m\,\mathrm{diag}A_m\bigr)\boldsymbol{\delta}_m=-\mathbf{g}_m$\nllabel{alg:lm:line:solve}\tcp*{Cholesky; non-PD $\Rightarrow$ fail}
  }
  \ParFor{active tree $m$}{
    \ParForR{data row $i$}{$e_{m,i}\leftarrow f_{T_m}(\mathbf{x}_i;\mathbf{c}_m+\boldsymbol{\delta}_m)-y_i$}
    $\ell_m\leftarrow\tfrac{1}{2}\textstyle\sum_i e_{m,i}^{2}$\tcp*{reduce}
  }
  \ForEach(\tcp*[f]{host code}){active tree $m$}{
    \eIf{$\ell_m\le\ell^{\star}_m$\nllabel{alg:lm:line:accept}}{
      $\mathbf{c}_m\leftarrow\mathbf{c}_m+\boldsymbol{\delta}_m$;\ \
      $\ell^{\star}_m\leftarrow\ell_m$;\ \ $\lambda_m\leftarrow0.1\,\lambda_m$\;
    }{
      $\lambda_m\leftarrow10\,\lambda_m$\;
    }
    status$_m\leftarrow$ converged / failed / active\;
  }
  \ParFor{active tree $m$}{refresh $\mathbf{r}_m$\nllabel{alg:lm:line:refresh}\;}
}
\Return $\mathbf{c}_m$, status$_m$
\end{algorithm}
The constant values live outside the tree structure: the whole population's
constants sit in one flat vector, addressed by a per-tree offset plus a
within-tree index. The solver updates only this vector; the structure arrays
are uploaded once and never rewritten. Each tree in the batch is a four-field
record: node offset, node count, constant offset, and constant count $K_m$.
Trees of any shape are just different ranges inside the same arrays, and this
is what lets a heterogeneous population share one batched launch. In the
tree-walking stages (evaluation and Jacobian construction), one warp serves
one tree and its 32 lanes split the $N$ data rows; the small per-tree solve
runs as one thread per tree. Each tree allows at most 32 constants, a 64-deep
operand stack, and a 128-node tape for reverse-mode automatic differentiation
(AD), all compile-time limits that a rebuild can raise. The solver reads only this buffer format and does
not care which engine produced the trees; adding a new source needs only a
thin host-side converter into this layout.

Each LM iteration builds every tree's Jacobian $J_m\in\mathbb{R}^{N\times
K_m}$ of the residual $\mathbf{r}_m=f_{T_m}(X;\mathbf{c}_m)-\mathbf{y}_m$ at~\Lcref{alg:lm:line:jacobian}, one
row per data point and one column per constant, then solves the damped normal equations at~\Lcref{alg:lm:line:solve} and accepts or rejects the trial step at~\Lcref{alg:lm:line:accept}. How each row of $J_m$ is built decides how the build cost scales with the tree's constant count $K_m$. We build it in three ways,
compared in \Cref{tab:builders}: finite differences (FD), AD
in forward mode (fwd AD) and reverse mode (rev AD). Rev AD is backpropagation over the
expression tree and matches the shape of the problem: one scalar residual
out, $K_m$ constants in. 
One taped forward walk plus one backward walk delivers the whole row at a
cost independent of $K_m$, the cheap-gradient property of reverse
mode~\cite{Griewank2008EvalDeriv}. The tape records at most two local
partials per node in per-lane scratch. The backward multiply uses a zero
guard: the product is taken as zero when either factor is exactly zero,
preventing a $0\times\infty$ at a singular point from producing a NaN that
poisons the whole column. This independence from $K_m$ applies only to the
Jacobian construction. Under all three modes, forming $J_m^{\top}J_m$ costs
$O(K_m^{2}N)$, and the per-tree Cholesky factorization costs $O(K_m^{3})$.
One LM iteration requires eight population-wide launches under all three
modes, independent of $M$ and of the trees' $K_m$. The Jacobian, residuals,
and normal-equation matrices remain on the GPU; each iteration, the host
exchanges only small per-tree state, and nothing that scales with $N$
crosses the bus.

\begin{table}[t]
\centering
\caption{Jacobian construction modes. Walks are per data row; build cost is
per tree per LM iteration.}
\label{tab:builders}
\scriptsize
\setlength{\tabcolsep}{2pt}
\begin{tabular}{llccc}
\toprule
mode & row construction & walks/row & build cost & per-lane extra \\
\midrule
FD & perturb one constant, re-walk & $K_m$ & $O(K_m N n_m)$ & 128\,B \\
fwd AD & 8 tangents/walk & $\lceil K_m/8\rceil$ & $O(\lceil K_m/8\rceil N n_m)$ & 2.0\,KB \\
rev AD & taped fwd + bwd sweep & $2$ & $O(N n_m)$ & 1.25\,KB \\
\bottomrule
\end{tabular}
\end{table}

\subsection{Double-Precision Delivery Guard}
\label{subsec:guard}

To retain the speed of the single-precision LM loop while protecting solution quality, we add a double-precision delivery guard. The numerical hot path runs in single precision (fp32), compiled under \texttt{-{}-use\_fast\_math}; trading numerical precision for throughput is common practice in high-performance GPU kernels~\cite{Fu2026Im2winMultiPrecision}. Near singularities, the reciprocal and square-root approximations of the fast-math can yield a residual that is finite but wrong. This error can fool the accept/reject decision: a step that looks like an improvement in single precision but turns out worse in double precision. Classical treatments of optimization under inexact arithmetic adjust the evaluation precision across iterations~\cite{Clancy2022TROPHY,GrattonToint2018}. We only need to certify the final result, not every iteration. We therefore perform the check at the delivery boundary. After the loop ends, every tree's loss is recomputed in double precision at both the delivered constants and the initial ones. Any tree whose delivered loss is non-finite or worse than its initial loss is rolled back to $\mathbf{c}^{0}_m$, its reported status and loss corrected to match. This check costs two double-precision evaluations per tree, done once after the loop, not every iteration. In double precision, delivered constants are never worse than initial ones, but the inner solve itself still runs in single precision.

\subsection{EvoGP Integration}
\label{sec:integration}

We integrate our CO implementation into EvoGP. Stock EvoGP samples constants from a fixed table or a bounded range but offers no mechanism for fitting them to the data; continuous inner constants — a frequency inside a sine, a decay rate inside an exponential — are therefore difficult to recover by sampling alone. Our integration closes this gap: before selection evaluates fitness, every candidate tree's constants are fit to the data by our solver.
The solver is loaded in-process by the EvoGP engine. Each generation, the constants from every tree are handed to the solver in one batched fit, and the returned values are written back; any tree whose fit does not improve keeps its inherited constants. The fit need not run every generation — invoking it every few generations preserves recovery quality at a fraction of the cost. Section~\ref{subsec:operon} quantifies the end-to-end improvement.


\section{Experimental Evaluation}


\subsection{Experimental Setup}
\label{subsec:setup}

\textbf{Hardware Platform} All GPU measurements use one NVIDIA A100-SXM4-80GB (compute capability 8.0, 80 GB HBM2e) hosted in a shared dual-socket AMD EPYC 7763 server ($2{\times}64$ cores). The GPU is held exclusively during each run, with its SM clock locked at 1410 MHz via \texttt{nvidia-smi -lgc} and ECC enabled. CPU baselines run as 64 worker processes pinned to the 64 physical cores of one socket; spanning both sockets measured 38\% lower Operon throughput under cross-NUMA memory-bandwidth contention.

\textbf{Software Environment} Kernels are built with CUDA 12.4. Library versions: Python 3.12, NumPy 2.1.3, SciPy 1.17.1~\cite{Virtanen2020SciPy,scipygit}, and pyoperon 0.6.1 (Operon rev. 5a1c937, single-precision release build)~\cite{operongit}; the end-to-end integration study additionally uses PyTorch 2.11 and EvoGP 0.1.0~\cite{evogpgit}. All GPU kernels are compiled with \texttt{-{}-use\_fast\_math}; the fast-math safety implications are addressed by the double-precision delivery guard described in Section~\ref{subsec:guard}.

\textbf{Measurement Methodology} Kernel time is measured on-device with CUDA events around the optimization loop itself, excluding the one-time ${\sim}280$~ms binary start-up (context creation and population load). Every configuration runs with three population seeds and three timed repetitions each, and we report medians. Headline throughput is quoted only from the occupancy-saturated regime (loop time ${\ge}145$~ms), where the median-of-repetitions throughput under locked clocks is reproducible within 0.5\%. No quality number in this section relies on the solver's own reporting: the constants delivered by every run are re-scored independently in fp64 on the host.

\subsection{Workload and Compared Implementations}
\label{subsec:workload}

We evaluate the constant-optimization step under controlled conditions. Its workload is driven primarily by the population statistics, specifically its tree shapes, depths, and constant counts, rather than by the target equation alone. Standard benchmark equations often contain only a few outer coefficients and therefore do not represent the many inner constants that arise during evolutionary search. Each benchmark therefore uses a synthetic population whose distribution of tree shapes, depths, and constant counts is calibrated to snapshots from real EvoGP runs.
 
We consider three workload regimes: early-generation populations of compact trees, late-generation populations dominated by bloat, and populations with many inner constants (per-tree constant counts average $1.2$--$6.9$ and never exceed $14$).
Targets are also synthetic: each tree's output at reference constants with $1\%$ Gaussian noise added. This yields a known near-global optimum for every instance, enabling direct assessment of solution quality and identical problem setups on CPU and GPU for fair throughput comparisons.
Populations range from
$10^{3}$ to $2.56{\times}10^{5}$ trees, with $100$ to $10{,}000$ data
points per tree.
All methods receive the same trees, data, and initial constants. 

We
compare three implementations of the same second-order method: our
solver (run with each of its three Jacobian modes), Operon as the
throughput baseline, and SciPy's \texttt{least\_squares} in \texttt{lm}
mode as a double-precision quality reference and additional
CPU baseline. All three implement LM updates from
the same MINPACK family. Both Operon and our solver construct the Jacobian
via reverse mode. The comparison therefore isolates execution
strategy: CPU methods optimize one tree per worker, whereas
our solver advances the entire population with a fixed number of
population-wide launches per iteration. We omit PySR for the same reason:
it optimizes constants with BFGS, not LM, so including
it would change the optimizer along with the execution strategy.

\begin{figure}[t]
\centering
\includegraphics[width=\columnwidth]{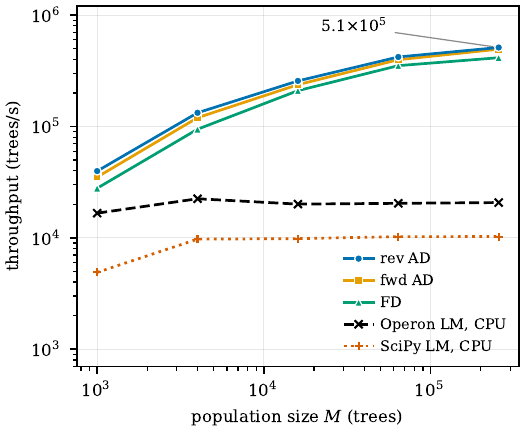}
\caption{Constant-optimization throughput versus population size $M$ at
$N{=}100$ (early-generation workload): the three Jacobian modes and the Operon
and SciPy CPU baselines. Axes are logarithmic.}
\label{fig:throughput}
\end{figure}

\subsection{Overall Throughput and Scaling Results}
\label{subsec:scaling}

Throughput is reported as optimized trees per second (trees/s), i.e., population size divided by optimization-loop wall time. \Cref{fig:throughput} shows the population-size scaling by sweeping $M$ with a fixed $N{=}100$ on the early-generation workload;
the same GPU scaling trend is observed across the remaining workload regimes.
Throughput climbs from $39{,}717$ trees/s at $M{=}10^{3}$ to $511{,}308$ trees/s at $M{=}256{,}000$, a $12.9\times$ rise; growth slows with $M$, and the curve is still rising at the largest population measured. 
This scaling is the payoff of population-wide batching: the fixed cost of a launch does not depend on how many trees it covers, so a small population leaves the device undersubscribed and the fixed cost
dominates, while a large population amortizes it. 

All three Jacobian modes exhibit the same scaling trend, and the early-generation workload is representative rather than easy: across the three structural regimes a population spans,
peak throughput ranges from $2.8{\times}10^{5}$ trees/s on the
constant-heavy regime to $6.1{\times}10^{5}$ on late-generation bloat.

Beyond population size, throughput also depends on the number of data points per tree. As $N$ grows from
$100$ to $10^{4}$ at $M{=}16{,}000$, per-tree work rises $100\times$ while
tree throughput falls only about $21\times$ (from $256{,}246$ to $12{,}340$
trees/s); the point rate meanwhile rises from $2.6{\times}10^{7}$ to
$1.2{\times}10^{8}$ points/s and has not leveled off, so larger $N$ keeps
the device fuller even as fewer trees finish per second.

Across the measured grid of three workloads, five population sizes, and
three data-point counts, rev AD is the fastest overall: its median
advantage is $1.51\times$ over FD and $1.11\times$ over fwd AD. Cell by
cell, the reverse-to-forward throughput ratio ranges from $0.95$ to
$1.43$: fwd AD wins narrowly where constants are few, and rev AD pulls
ahead as $K_m$ grows, as the walk counts of \Cref{tab:builders}
predict. 
So, rev AD is the default.

\subsection{Bottleneck Explanation and In-Loop Cost}
\label{subsec:bottleneck}

The optimization loop is neither compute-bound nor bandwidth-bound. Across
the Jacobian and evaluation kernels, Nsight Compute speed-of-light profiles
show FMA-pipe utilization at or below $15\%$ ($11\%$ for the Jacobian
kernel) and DRAM utilization below $23\%$ and mostly under $10\%$. 
An instruction-roofline view places the kernels between $24\%$ and $55\%$ of
the device's instruction-issue ceiling, 
and the dominant stall reasons are
dependency waits and memory-latency scoreboard stalls. The loop is
therefore bound by instruction issue and on-chip latency: its work consists
of many short dependent walks over irregular trees, not dense arithmetic.
Population-wide batching supplies the throughput here: it keeps enough
independent trees in flight to hide the per-lane latency of these dependent walks.

In stage-level timing, building the Jacobian is the largest stage under FD, $27$--$60\%$ of loop time (peaking at $M{=}16{,}000$, $N{=}1{,}000$); rev AD holds the same stage to $12$--$28\%$, consistent with the walk counts of the two modes. 
Normal-equation assembly accounts for $6$--$15\%$ of the loop time, and the
per-tree Cholesky factorization only $2$--$5\%$.
The per-iteration host round trip (constants up; step and status down) takes $5$--$11\%$ of the loop at $N{=}1{,}000$ and $11$--$29\%$ at $N{=}100$, part of the fixed cost that dominates the small-$N$ end.

The compact, unpadded layout is the loop's other lever, echoing how memory-efficient data layouts govern throughput in other GPU kernels~\cite{Lu2023Im2winGPU}. Fixed-shape GPU
frameworks pad every tree to a run-level node cap; padding to our
$128$-node cap runs the solver $4.45{\times}$ slower on the early-generation
workload at $M{=}16{,}000$, $N{=}1{,}000$, on identical inputs through the same binary, returning bit-identical results, so the
difference is layout alone. EvoGP's tighter configured length ($64$ nodes) pays
less but cannot escape the tax: the gap between a worst-case cap and
populations that here average $12$--$28$ nodes. Within our cap a growing tree is a longer range, while a fixed-length layout must re-pad the whole population
and cannot represent trees that outgrow its cap at all.

\subsection{Quality, Throughput, and In-Loop Integration}
\label{subsec:operon}

We measure quality on a fixed population of $1{,}000$ EvoGP trees, solved independently by the three GPU Jacobian modes and by SciPy's fp64 using the same data and initial constants.
The GPU modes produce similar rankings: Spearman correlations are $0.959$ between rev AD and FD and $0.995$ between the two AD modes, with $95.5\%$ overlap in their top-$10\%$ sets. Jacobian mode is therefore primarily a throughput choice.
Compared with the SciPy reference, $86.6$--$93.8\%$ of converged trees fall within $1.05\times$ of the reference loss (Table~\ref{tab:accuracy}).
Since evolutionary selection depends mainly on fitness ranks, these small deviations are unlikely to materially affect which high-fitness trees survive. 
Table~\ref{tab:accuracy} also reports Operon's results for reference. Operon converges in a median of five iterations and improves $45\%$ of the trees. On this near-optimal synthetic fixture, the loss-down metric mainly reflects stopping behavior near the fp32 precision floor rather than solver quality.

Having established comparable solution quality, we compare throughput at a representative saturated configuration. 
On the early-generation workload with $M{=}16{,}000$ and $N{=}1{,}000$, one A100 delivers $98{,}154$ trees/s compared with $9{,}959$
trees/s for Operon on one EPYC~7763 socket and $5{,}310$ trees/s for the
SciPy reference, corresponding to speedups of about $9.9\times$ and $18\times$. The advantage over Operon depends on the number of data points per tree: at fixed $M{=}16{,}000$, it narrows from $12.8\times$ at $N{=}100$ to $2.7\times$ at $N{=}10^{4}$. This narrowing does not indicate reduced GPU efficiency: over the same range, our point-evaluation throughput rises from $25.6$ to $123.4$ million points/s. Instead, Operon benefits more from amortizing fixed per-tree overheads as $N$ increases.

\begin{table}[t]
\centering
\caption{Share of trees whose delivered loss lands within the stated factor
of the SciPy fp64 reference, among trees where both solvers converge; loss
down: share of trees with constants whose delivered loss improves on the
initial value.}
\label{tab:accuracy}
\scriptsize
\setlength{\tabcolsep}{4pt}
\begin{tabular}{lcccc}
\toprule
 & \multicolumn{3}{c}{within factor of reference} & \\
\cmidrule(lr){2-4}
mode & $1.05\times$ & $2\times$ & $10\times$ & loss down \\
\midrule
FD & 93.8\% & 95.7\% & 99.7\% & 94.0\% \\
fwd AD & 86.6\% & 90.0\% & 100.0\% & 93.3\% \\
rev AD & 86.6\% & 90.0\% & 100.0\% & 93.3\% \\
\midrule
Operon & 73.9\% & 75.0\% & 75.9\% & 45.0\% \\
\bottomrule
\end{tabular}
\end{table}

The results so far establish standalone quality and throughput. We integrate the GPU LM solver directly into EvoGP without changing the search. We construct $18$ test problems whose true equations each place
a constant inside a nonlinear function. We judge recovery by symbolic equivalence to the true equation rather than $R^2$, which reflects only numerical closeness of fit, not whether the structure is recovered. 
Stock EvoGP can reach $R^2 > 0.999$ with bloated expressions that do not recover the real equation.
With constant optimization, the same search recovers $10$ of the $18$ problems;
without it, none (exact McNemar test over problems, $p{=}2.0{\times}10^{-3}$). The cost
of this step can be amortized further: constant optimization need not run every
generation, and applying it every fifth generation recovers almost as many
equations as every generation ($p{=}0.73$). This section implements the
simplest in-process integration; jointly optimizing cadence with the evolutionary search remains future work.

\section{Conclusion}
We presented a GPU-resident, batched Levenberg--Marquardt solver that makes
constant optimization practical inside GPU-accelerated tree-based genetic programming. On early-generation populations, the solver sustains up to $5.1\times 10^{5}$~trees/s
on an A100; at a saturated reference cell it delivers about $9.9\times$
the throughput of Operon on a 64-core EPYC~7763, matching fp64-reference
quality. Integrated in-process into EvoGP, the search
recovers the governing equations on $10$ of $18$ inner-constant problems; stock EvoGP recovers none. 
These results remove a key CPU-side bottleneck in GPU-accelerated tree-based genetic programming, enabling structure search and nonlinear constant fitting to execute together on GPU.

\section*{Acknowledgment}

This work was partially supported by the National Natural Science Foundation of China under Grant No. 12572266. The authors gratefully acknowledge the support provided for this research. This work is not related to Xu T. Liu's position at Amazon. 

\bibliographystyle{IEEEtran}
\bibliography{bibliography}

\end{document}